\documentclass[10pt,twocolumn,letterpaper]{article}

\usepackage{cvpr}
\usepackage{color}
\usepackage{algorithm}
\usepackage{algpseudocode}
\usepackage{float}
\usepackage{multirow}
\usepackage{makecell}
\usepackage{amsmath}
\usepackage{graphicx}
\usepackage{multirow}
\usepackage{colortbl}
\usepackage{amssymb}

\usepackage[dvipsnames]{xcolor}

\definecolor{cvprblue}{rgb}{0.21,0.49,0.74}
\usepackage[pagebackref,breaklinks,colorlinks,citecolor=cvprblue]{hyperref}

\def\paperID{13583} 
\def\confName{CVPR}
\def\confYear{2025}

\title{LaVida Drive: Vision-Text Interaction VLM for Autonomous Driving with Token Selection, Recovery and Enhancement}

\author{
    Siwen Jiao$^{1,3\textsuperscript{*}}$, Yangyi Fang$^{2\textsuperscript{*}}$,Baoyun Peng$^{4\textsuperscript{*†}}$,Wangqun Chen$^{4}$,Bharadwaj Veeravalli$^{1}$,Xulei Yang$^{3}$
\\
    $^1$National University of Singapore \\
    $^2$Tsinghua University \\
    $^3$Agency for Science, Technology and Research, Singapore \\
    $^4$Advanced Institute of Big Data, Beijing \\
}

\begin{document}
\maketitle
\renewcommand{\thefootnote}{\fnsymbol{footnote}}
\footnotetext[1]{Equal contribution.}
\footnotetext[2]{Corresponding author}
\renewcommand{\thefootnote}{\arabic{footnote}}
\begin{abstract}
Recent advancements in Visual Language Models (VLMs) have made them crucial for visual question answering (VQA) in autonomous driving, enabling natural human-vehicle interactions. 
However, existing methods often struggle in dynamic driving environments, as they usually focus on static images or videos and rely on downsampling to manage computational costs. This results in the loss of critical details and the difficulty in effectively integrating spatial and temporal information, undermining fine-grained perception and temporal coherence essential for effective decision-making.
To tackle these challenges, we introduce LaVida Drive, a novel and efficient VQA framework for autonomous driving. LaVida Drive seamlessly integrates temporal data while maintaining high-resolution inputs for detailed visual perception. It optimizes spatial processing by retaining high-resolution data for intricate details and using lower-resolution inputs for temporal analysis to focus on motion-related features, thereby boosting computational efficiency. Our method achieves an impressive 168-fold token compression while attaining optimal performance, a significant improvement over traditional approaches.
The core of LaVida Drive consists of two modules: the \textit{Query-aware Token Selection} module and the \textit{Spatial-Temporal Token Recovery and Enhancement} module. The former dynamically selects the most relevant visual tokens based on semantic alignment with the input query, reducing the token count from high-resolution spatial input. The latter ensures smooth and coherent interactions between spatial and temporal information, preserving contextual continuity across frames.
Extensive experiments on various autonomous driving question-answering benchmarks show that LaVida Drive significantly reduces visual tokens, enhances efficiency, and improves overall performance.

\end{abstract}    
\section{Introduction}
\begin{figure}[ht]
    \centering
    \includegraphics[width=1.00\linewidth]{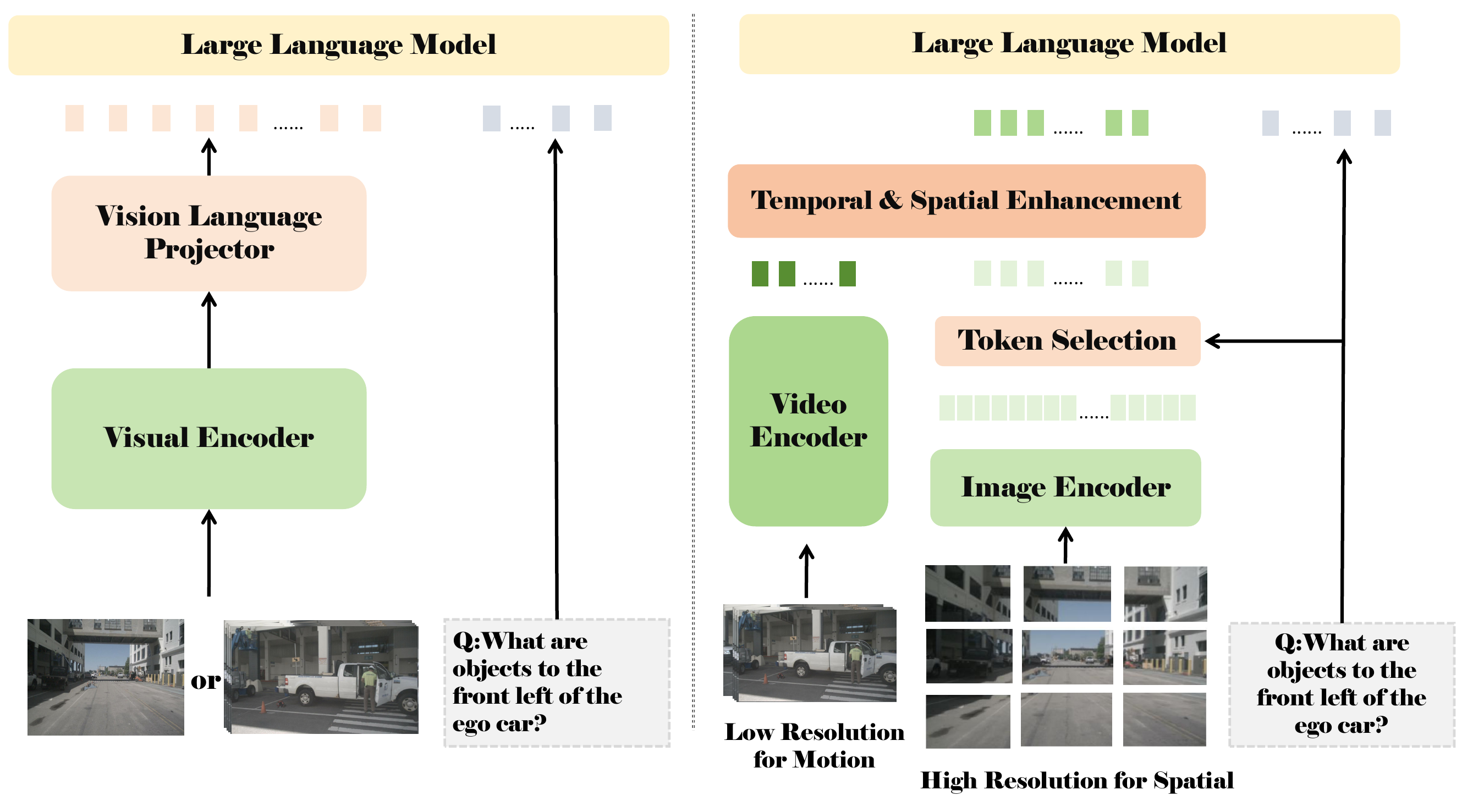}
    \caption{Comparison between LaVida Drive and existing VLM-based autonomous driving methods. (Left): Traditional methods use a VIT projector with simple monocular input and downsampling. (Right): LaVida Drive efficiently captures high-resolution details while maintaining motion perception, replacing the projector with enhanced modules that leverage encoder-level tokens for spatial and temporal enhancement.
    } 
    \label{fig:1} 
\end{figure}

Recent advancements in large-scale pre-training have positioned VLMs as pivotal for VQA in autonomous driving, enabling intuitive human-vehicle interactions through natural language \cite{liao2024vlm2scene, qian2024nuscenes, chen2024driving, sima2023drivelm, wen2023road, nie2025reason2drive}. VLMs facilitate the seamless integration of visual and linguistic information, allowing vehicles to comprehend and respond to complex queries in real-time, interpreting dynamic environments quickly and significantly improving the overall performance and reliability of the system \cite{nie2025reason2drive, sima2023drivelm, wang2024driving}.

Despite significant advancements, existing approaches predominantly solely focus on static images or video and rely on low-resolution inputs to manage computational costs, leading to the loss of critical high-resolution details and the difficulty in effectively integrating spatial and temporal information\cite{huang2023language,zhang2021vinvl,tian2024drivevlm,xu2024drivegpt4}. 
This is particularly problematic in dynamic driving environments, where downsampling impairs fine-grained perception and temporal coherence, hindering effective decision-making \cite{nie2025reason2drive, wang2024driving,tian2024drivevlm,yang2024generalized}. Balancing efficiency and accuracy in high-resolution, multi-frame settings for both static perception and motion detection significantly increases inference costs, posing a major challenge in VLM development \cite{wang2024driving, wen2023road, yang2024generalized, sha2023languagempc}.

To address these challenges, we propose LaVida Drive, an innovative VQA framework designed to support fine-grained perception of high-resolution visual inputs in dynamic driving environments while integrating temporal information. Specifically, for spatial processing, the framework retains high-resolution inputs to capture rich details and uses lower resolution for temporal processing on motion-related features, thereby reducing computational load without compromising visual accuracy. However, maintaining high-resolution spatial inputs across multiple viewpoints significantly increases the number of tokens, leading to substantial inference overhead in VLMs.
To handle this, we introduce the \textit{Query-aware Token Selection} mechanism, which dynamically selects visual tokens highly relevant to the input query based on semantic content, enabling adaptive token filtering and significantly easing the burden of computation \cite{nie2025reason2drive, yang2024generalized}. Since token selection would disrupt spatial coherence and damage the contextual relationships between tokens, we introduce a \textit{Spatial-temporal Token Enhancement} module to ensure coherence across different spatial and temporal contexts by using cross-attention mechanisms for consistent information flow across frames, achieving smooth and coherent multi-frame information transfer.

We validate LaVida Drive across multiple autonomous driving VQA benchmarks, showing significant enhancements in image-text alignment and multi-modal information processing. Our model reduces visual tokens by 50\% to 84\%, improving inference efficiency while maintaining performance. Key contributions include:

\begin{itemize}
\item Propose a novel and efficient VQA framework that seamlessly integrates temporal data into high-resolution spatial inputs, enhancing computational efficiency and detailed visual perception.
\item Propose a novel query-aware token selection mechanism that dynamically extracts key information for question answering, demonstrating its effectiveness in balancing computation cost and performance.
\item Propose a token enhancement mechanism integrating multi-modal and multi-scale information, ensuring smooth and coherent interactions between spatial and temporal information and preserving contextual continuity across multiple frames.
\end{itemize}

\section{Related Works}
\label{sec:related_works}

Recent advancements in autonomous driving have leveraged the intersection of vision and LLMs, driving improvements in both perception and decision-making capabilities. The literature can be categorized into two main areas: vision-based LLMs for autonomous driving and QA systems in autonomous driving.

\subsection{Vision-based LLMs for Autonomous Driving}
The integration of vision and language models has shown great promise in enhancing the perception capabilities of autonomous vehicles, enabling them to better understand and navigate complex driving environments. Early work in this domain includes CLIP-based methods~\cite{radford2021learning}, which pair visual representations with textual descriptions, enabling a richer understanding of the vehicle’s surroundings. Recent studies, such as those by~\cite{zhu2023vision} and~\cite{zeng2023multimodal}, have proposed large multi-modal models incorporating visual and textual inputs to support decision-making. These models benefit from pretraining on large-scale datasets and have shown improvements in tasks such as scene interpretation and predicting vehicle behaviours in dynamic traffic scenarios. 

The application of transformer-based models to vision-language fusion has also led to promising developments in autonomous driving. For instance, \cite{chen2022deep} developed a model that combines deep vision transformers with large-scale language models, which improves decision-making capabilities by enhancing the vehicle’s ability to generate complex driving plans. These models process real-time visual input while leveraging pre-trained knowledge to interpret high-level cues, such as road conditions and traffic regulations. Recently, \cite{zhang2024end} demonstrated the ability of multi-modal LLMs to engage in reasoning tasks in an end-to-end autonomous driving framework, allowing the vehicle to handle novel driving situations that require both visual and linguistic reasoning.

Furthermore, the advent of VLMs has opened new avenues for enhancing autonomous driving systems. For example, \cite{qian2024nuscenes} introduced NuScenes-QA, a benchmark for VQA in autonomous driving scenarios, which addresses the complexities of multi-modal data and real-time acquisition. Similarly, \cite{sima2023drivelm} proposed DriveLM, a VLM-based approach that integrates web-scale data to boost generalization and interactivity with human users. These advancements highlight the potential of VLMs to address the nuanced challenges of autonomous driving, such as understanding dynamic environments and making informed decisions in real time.

\subsection{Question Answering Systems for Autonomous Driving}

QA systems have played a crucial role in improving human-vehicle interaction and facilitating autonomous decision-making. In autonomous driving, these systems help vehicles process natural language queries and provide context-aware answers based on visual inputs and pre-existing knowledge. For example, \cite{yang2022vision} developed a visual QA system that combines convolutional neural networks with language models to allow autonomous vehicles to answer questions about nearby objects and road conditions. This system allows passengers to ask real-time questions, and receive accurate, context-specific responses.

Further advances in contextual question answering~\cite{chen2023contextual} have significantly enhanced the vehicle’s ability to interpret complex driving scenarios. By utilizing multi-modal input, these systems provide more accurate answers to questions regarding traffic flow, pedestrian movement, or vehicle proximity. In addition, dialogue-based QA systems have gained traction in recent years, enabling more dynamic interactions between drivers and vehicles. For instance, \cite{li2024dialogue} introduced a conversational QA framework, where vehicles answer questions and can engage in multi-turn dialogues, adjusting their responses based on evolving traffic conditions and user preferences. This enables smoother communication between passengers and vehicles, improving the overall driving experience and safety. 

More recent work by \cite{wang2024hybrid} explores hybrid models that combine rule-based reasoning with large-scale language models, allowing vehicles to simulate human-like reasoning during real-time decision-making in complex environments more accurately. Their work focuses on providing accurate and safe driving suggestions during ambiguous driving situations, such as when encountering unforeseen obstacles or pedestrians. Additionally, the integration of LLMs into QA systems has shown significant promise. For example, \cite{chen2024driving} proposed a unique object-level multi-modal LLM architecture that merges vectorized numeric modalities with a pre-trained LLM to improve context understanding in driving situations. This approach not only enhances the interpretability of driving actions but also demonstrates the potential of LLM-based driving action generation compared with traditional behavioural cloning methods.

\section{Method}
\subsection{Architecture Overview}

\begin{figure*}[ht]
    \centering    
    \includegraphics[width=\textwidth, trim=0 0 850 0, clip]{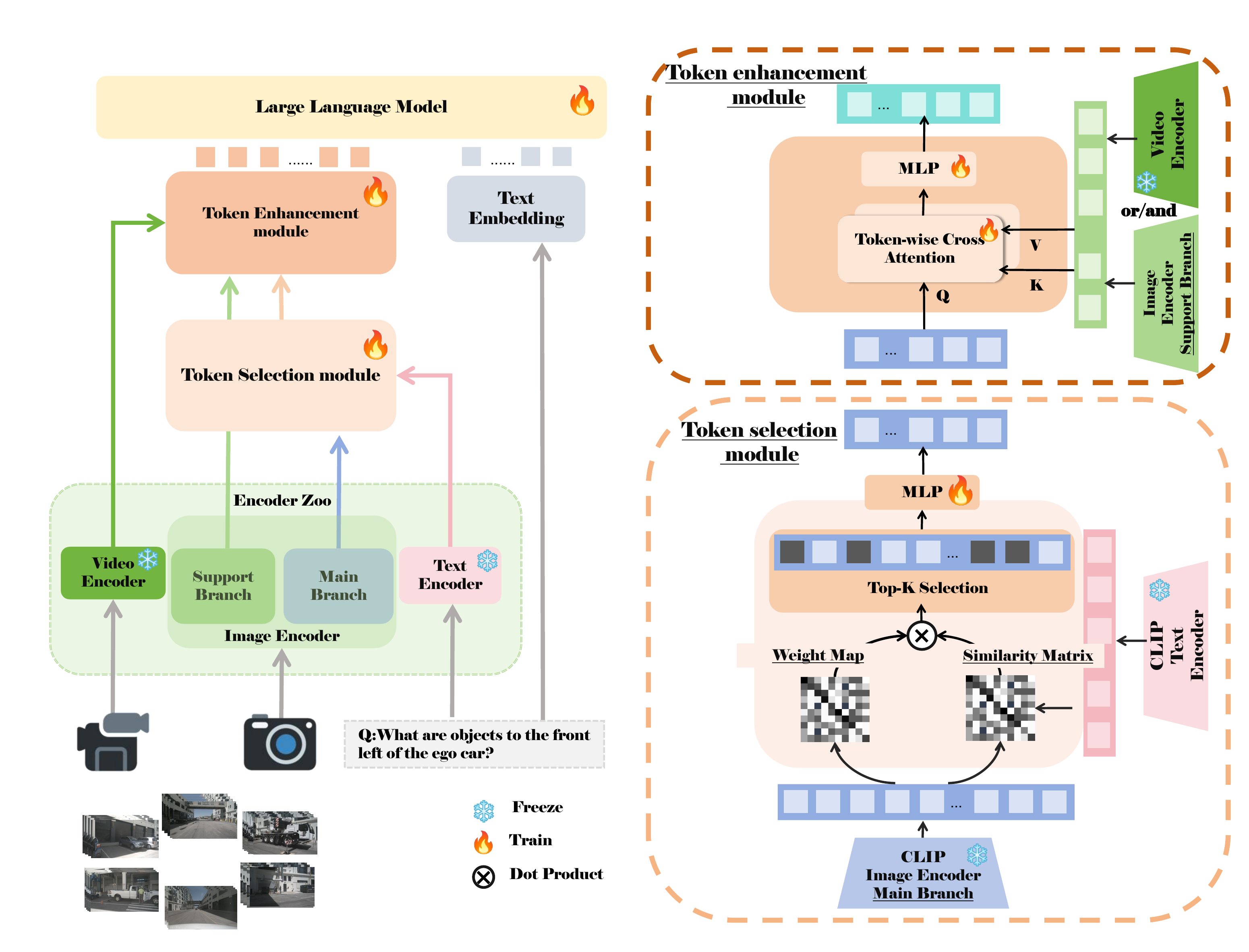} 
    \caption{Framework of LaVida Drive. High-resolution images are divided into $224\times 224$ patches and processed by the image encoder to extract semantic features. A token-level similarity matrix aligns image tokens with text tokens, enabling query-aware token selection. Selected tokens are enhanced through a token-wise attention mechanism in the enhancement module, utilizing auxiliary branches for restoration without increasing token count. The final tokens and original text embeddings are then fed into the large language model.}
    \label{fig:pipeline}  
\end{figure*}

As illustrated in Fig.~\ref{fig:pipeline}, the architecture of LaVida Drive comprises three core components: the Multi-modal Encoder Cluster, the Query-aware Token Selection Module, and the Spatial-temporal Token Enhancement Module. The model processes inputs from three modalities: image data from the autonomous vehicle's multiview cameras, video data, and natural language instructions provided by the user.
Different from previous approaches, we employ multiple encoders to handle various input modalities, forming a \textbf{Multi-modal Encoder Cluster} to better address the unique requirements of each data source. All encoders are frozen. Specifically, each encoder processes data in a predefined format:

\textbf{Text Encoder:} Our text encoder employs the CLIP text encoder, leveraging its powerful feature extraction capabilities obtained through large-scale text-image contrastive learning. For an input text sequence $\mathbf{X}_{\text{text}}$ containing $L_{\text{text}}$ tokens, the encoder processes each token embedding and maps the entire sequence to a semantic space. The output of the text encoder is a matrix of shape $L_{\text{text}} \times d$, represented as:
\begin{equation}
\mathbf{E}_{\text{text}} = \text{TextEncoder}(\mathbf{X}_{\text{text}}) \in \mathbb{R}^{L_{\text{text}} \times d}.
\end{equation}

\textbf{Image Encoder:} The image encoder also employs the CLIP visual encoder, with a base resolution of $224 \times 224$ pixels. This encoder efficiently maps visual data into a rich semantic space and is divided into a \textit{main branch} and a \textit{support branch}, each optimized for different aspects of image representation.

\textit{Image Encoder Main Branch:} For the main branch, an input image $\mathbf{X}_{\text{img}}$ of size $H \times W \times C$ is first divided into $N = \frac{H \times W}{P^2}$ patches of size $P \times P \times C$. Each patch is flattened into a vector of length $P^2 \times C$, forming a patch sequence of dimension $N \times (P^2 \times C)$. The main branch generates embeddings of shape $L_{\text{main}} \times d$ from the penultimate layer of the CLIP visual encoder, represented as:
\begin{equation}
\mathbf{E}_{\text{main}} = \text{MainBranchEncoder}(\mathbf{X}_{\text{img}}) \in \mathbb{R}^{L_{\text{main}} \times d}.
\end{equation}

\textit{Image Encoder Support Branch:} To supplement the context loss due to patch segmentation in the main branch, the support branch directly processes the downsampled entire image $\mathbf{X}_{\text{img}}'$, of size $224 \times 224 \times C$. The support branch also generates embeddings of shape $L_{\text{support}} \times d$ from the penultimate layer of the CLIP visual encoder, represented as:
\begin{equation}
\mathbf{E}_{\text{support}} = \text{SupportBranchEncoder}(\mathbf{X}_{\text{img}}') \in \mathbb{R}^{L_{\text{support}} \times d}.
\end{equation}

\textbf{Video Encoder:} The video encoder is based on the TimeSformer model, which performs temporal modelling on frame sequences. Given an input sequence $\mathbf{X}_{\text{temporal}}$ of $T$ frames, where each frame has a spatial dimension of $H \times W \times C$, the encoder captures inter-frame dependencies to generate temporal representations. The output is an embedding sequence of size $T \times d$, represented as:
\begin{equation}
\mathbf{E}_{\text{temporal}} = \text{VideoEncoder}(\mathbf{X}_{\text{temporal}}) \in \mathbb{R}^{T \times d}.
\end{equation}

Next, we employ the Query-aware Token Selection Module, which processes tokens output by the image encoder and text encoder to generate a token-level similarity matrix \( S \in \mathbb{R}^{m \times n} \), where \( m \) denotes the number of image tokens and \( n \) denotes the number of text tokens. By leveraging semantic similarity in space, this module identifies the most relevant visual tokens to the user's query, thereby reducing the number of visual tokens while retaining high-quality tokens.
Finally, the Spatial-temporal Token Enhancement Module utilizes the video encoder’s output $V_{3D} \in \mathbb{R}^{S_{3D} \times 768}$ and the multi-frame auxiliary information from the image encoder $V_{2D_{multi-frame}} \in \mathbb{R}^{S_{2D} \times 768}$ to recover and enhance tokens through a cross-attention mechanism. This module’s purpose is to restore context lost in token selection and aggregate temporal information without increasing the number of additional tokens, as described further in Section 3.3.
\begin{figure}
    \centering
    \includegraphics[width=0.9\linewidth]{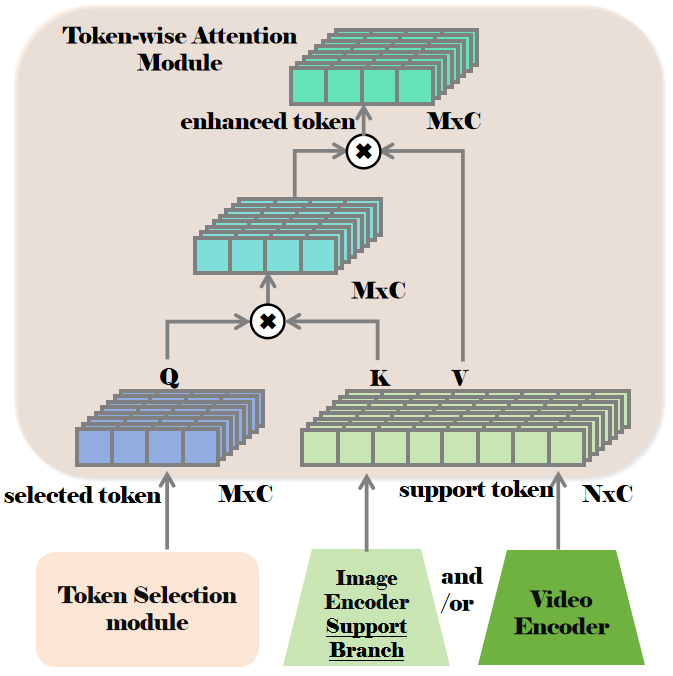}
    \caption{The architecture of the Token-wise Attention module. We leverage the outputs from the support branches of the image encoder and video encoder, applying cross-attention to recover/enhance the selected tokens while maintaining consistent output dimensions according to the rules of attention computation.}
    \label{fig:3} 
\end{figure}
\subsection{Query-aware Token Selection}
The Query-aware Token Selection module filters out the most relevant and important tokens based on image and text tokens and further compresses these tokens for efficient representation.
In the section of the \textbf{Multi-modal Encoder Cluster}, we denote the main branch output from the image encoder as $\mathbf{I}$ and the output from the text encoder as $\mathbf{T}$.Given that the CLIP\cite{Radford2021LearningTV} model was pre-trained on a large image-text dataset, we assume that image and text tokens are mapped into the same semantic space.To compute token-wise similarity, we omit the 0-th dimension from the embeddings, as it represents the overall semantic features. 

To align the image embeddings with the text embeddings in the same semantic space, we apply a multi-layer perceptron (MLP) to the image tokens. This transformation produces the aligned image representation, denoted as $\mathbf{I}'$. The aligned image embeddings $\mathbf{I}'$ can then be used for similarity computation with the text embeddings $\mathbf{T}$. Therefore, we calculate the cosine similarity to obtain a token-wise similarity matrix $\mathbf{S}$, measuring the semantic similarity between each image and text token.
\begin{equation}
s(\mathbf{I}', \mathbf{T}) = \frac{\mathbf{I}' \cdot \mathbf{T}^{\top}}{\|\mathbf{I}'\| \|\mathbf{T}\|}.
\end{equation}

Next, based on the similarity between each image token and text token, we calculate the normalized similarity matrix as follows:

\begin{equation}
p_{i}^{\text{img}}(x) = \frac{\exp(s(\mathbf{I}', \mathbf{T}_{i})/\tau)}{\sum_{j=1}^{N} \exp(s(\mathbf{I}', \mathbf{T}_{j})/\tau)}.
\end{equation}

This is then used to select the most relevant $\mathbf{k}$ image tokens, where $\mathbf{k}$ represents the sampling threshold. A higher threshold selects more visual tokens, allowing the quantity of visual tokens to be controlled by adjusting the downsampling ratio $\mathbf{k}$. The experimental section compares performance under different thresholds, highlighting their impact as a proportion of the total token count. The steps are outlined in the following algorithm:

\begin{algorithm}
\caption{Query-aware Token Selection}

\textbf{Input:} Image tokens $\mathbf{I}$, Text tokens $\mathbf{T}$, temperature parameter $\tau$, learnable parameter $\alpha$, number of top-k tokens to select $k$ \\
\textbf{Output:} Selected top-k image tokens $\mathbf{T}_{\text{top-k}}$

\begin{itemize}
    \item \textbf{Align Image and Text Embeddings:}\\ 
    $\mathbf{I}' \gets \text{MLP}(\mathbf{I})$

    \item \textbf{Compute Cosine Similarity:} \\
    $s(\mathbf{I}', \mathbf{T}) \gets \frac{\mathbf{I}' \cdot \mathbf{T}^{\top}}{\|\mathbf{I}'\| \|\mathbf{T}\|}$

    \item \textbf{Normalize Similarity:} \\
    $p_{i}^{\text{img}}(x) \gets \frac{\exp(s(\mathbf{I}', \mathbf{T}_{i})/\tau)}{\sum_{j=1}^{N} \exp(s(\mathbf{I}', \mathbf{T}_{j})/\tau)}$

    \item \textbf{Compute Similarity Scores Matrix:}\\ 
    $\mathbf{S}_{\text{sum}} \gets \sum_{k=1}^{K} p_{i}^{\text{img}}(x)$

    \item \textbf{Compute Token Weights Matrix:}\\
    $\mathbf{W} \gets \sum_{i} \mathbf{I}'_{i}$

    \item \textbf{Compute Selection Map:}\\
    $\mathbf{M} \gets (1 - \alpha) \cdot \mathbf{S}_{\text{sum}} + \alpha \cdot \mathbf{W}$

    \item \textbf{Select Top-K Tokens:} \\
    $\mathbf{T}_{\text{top-k}} \gets \mathbf{I}'[\text{TopK}(\mathbf{M}, k)]$

    \item \textbf{Return:}
    $\mathbf{T}_{\text{top-k}}$
\end{itemize}

\end{algorithm}

To further compress the tokens, we employ an MLP for information aggregation, generating a query-aware compact token representation $\mathbf{{T}_{\text{select}}}$. In the experimental section, we demonstrate the model's performance under various combinations of selection and MLP compression factors at a fixed overall compression ratio. The results indicate that balancing the selection factor with the MLP compression rate can achieve higher model performance while preserving a minimal number of tokens.

\subsection{Spatial-temporal Token Enhancement}

The Spatial-temporal Token Enhancement module is designed to address issues of contextual disruption and high computational overhead when dealing with multi-frame data. This module includes a general-purpose \textbf{Token-wise Attention Module}, which is then specifically applied in two configurations: spatial and temporal enhancements.

First, we introduce the \textbf{Token-wise Attention Module}, which forms the foundation for both spatial and temporal enhancements. As shown in Fig. \ref{fig:3}, this module enhances token context by enabling interactions between the tokens selected by the Query-aware Token Selection module and the context tokens from the image or video encoder. The attention mechanism is defined as follows:

\begin{equation}
\text{Att}_{\text{token-wise}}(Q, K, V) = \text{Softmax}\left(\frac{QK^T}{\sqrt{d_k}}\right)V,
\end{equation}

where \( Q \) (Query) represents token representations from the Query-aware Token Selection module, while \( K \) (Key) and \( V \) (Value) are derived from the encoder outputs. This mechanism allows each query token to absorb relevant information from context tokens.

With the Token-wise Attention Module established, we now apply it in two configurations:

\begin{itemize}
    \item \textbf{Spatial Token Restoration}: Here, we use the output \( E_{\text{spatial}} \) of the image encoder as keys and values. By interacting with each token representation \( I' \), this configuration introduces spatial context, resulting in the spatially enhanced representation \( I_{\text{enhanced spatial}} \):
\begin{align}
    I_{\text{enhanced spatial}} &= \text{Att}_{\text{token-wise}}(Q = I', \\
    &\quad K = E_{\text{spatial}}, V = E_{\text{spatial}}) \notag.
\end{align}
    \item \textbf{Temporal Token Enhancement}: When handling multi-frame data, we use the output \( E_{\text{temporal}} \) of the video encoder as another set of keys and values. By interacting in parallel with \( I' \), this configuration incorporates temporal context and produces the temporally enhanced representation \( I_{\text{enhanced temporal}} \):
\begin{align}
    I_{\text{enhanced temporal}} &= \text{Att}_{\text{token-wise}}(Q = I', \\
    &\quad  K=E_{\text{temporal}}, V=E_{\text{temporal}}) \notag.
\end{align}

\end{itemize}

Finally, we combine the spatial and temporal enhanced representations through an MLP layer to obtain the final token representation:

\begin{equation}
I_{\text{final}} = \text{MLP}(I_{\text{enhanced spatial}} + I_{\text{enhanced temporal}}).
\end{equation}

In this way, the Spatial-temporal Token Enhancement module fully leverages both spatial and temporal information without increasing computational costs. The final output tokens are then concatenated with the query embedding for input into the Large Language Model.

\section{Experiments}
\label{sec:Experiments}
\begin{figure*}[ht]
    \centering
    \includegraphics[width=0.95\linewidth, trim=0 20pt 0 20pt, clip]{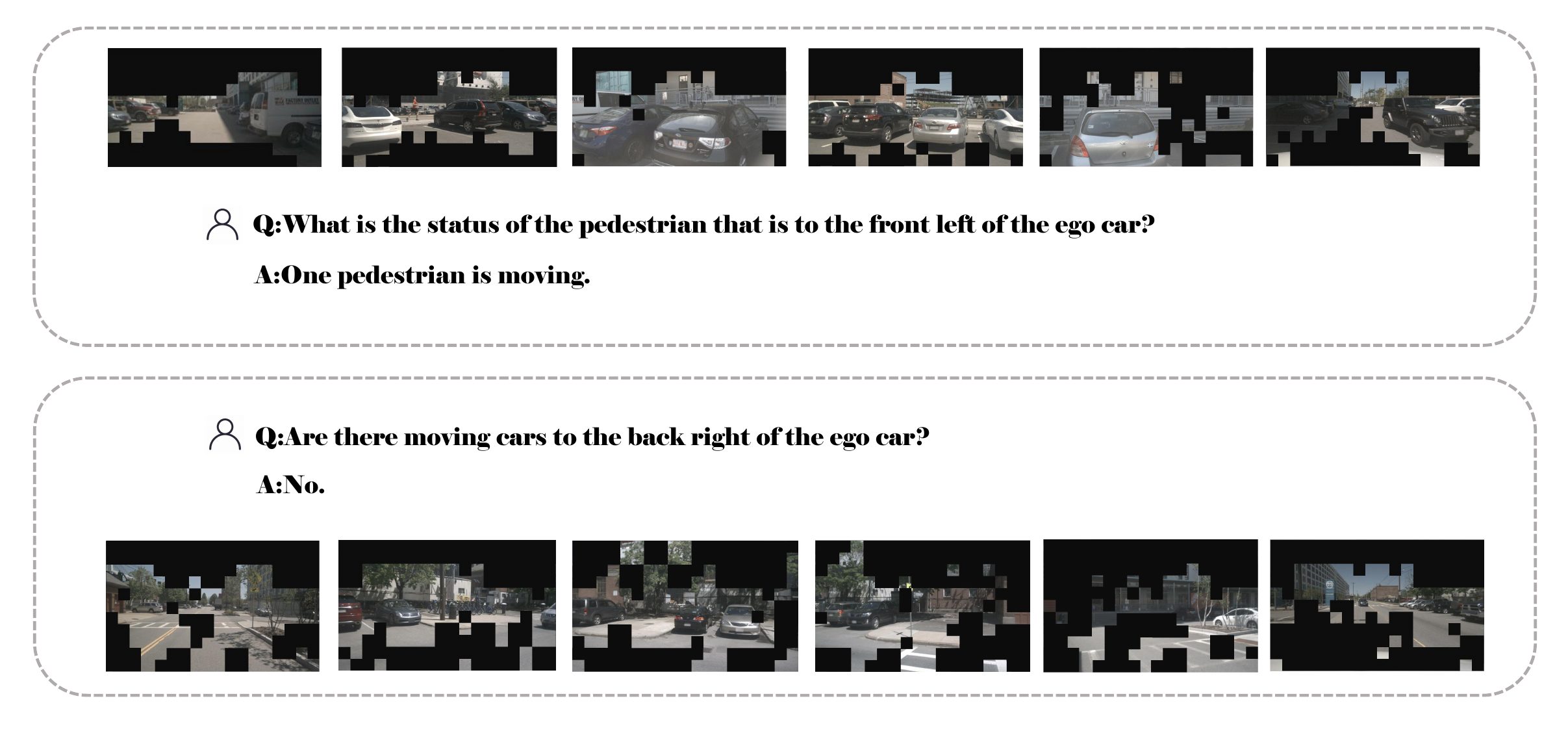} 
    \caption{Lavida Drive Response Example.
The areas not covered by the black mask represent the image patches corresponding to the retained tokens. This is provided as a simplified example for ease of visualization, and the actual results may differ.} 
    \label{fig:4} 
\end{figure*}

\begin{table*}[ht]
\caption{Performance Comparison on DriveLM and NusceneQA Datasets. Bold indicates the highest value, while an underline indicates the second-highest value.}
\centering
\setlength{\tabcolsep}{5mm}  
\resizebox{\textwidth}{!}{
\begin{tabular}{l|c|c|cccc}
\toprule
\textbf{Dataset} & \textbf{Method} & \textbf{Ref.} & \textbf{BLEU-4↑} & \textbf{METEOR↑} & \textbf{ROUGE-L↑} & \textbf{CIDEr↑} \\
\midrule
\multirow{4}{*}{DriveLM Dataset}  & EM-VLM4AD\textsubscript{Base}  & {CVPR’24} & 45.4 & 34.5 & 72.0 & 3.20  \\
                                  & EM-VLM4AD\textsubscript{Large} & {CVPR’24} & 40.1 & 34.3 & 70.7 & 3.10  \\
                                  & DriveLM-Agent                   & {ECCV’24} &\textbf{53.1} & 36.2 & 66.8 & 2.79 \\
                                  & LaVida Drive (Ours)             & {-}       & \underline{51.3} & \textbf{38.0} & \textbf{73.9} & \textbf{3.32} \\
\bottomrule
\end{tabular}
}
\vspace{0.5cm}

\resizebox{\textwidth}{!}{
\begin{tabular}{l|c|c|ccccc}
\toprule
\textbf{Dataset}   & \textbf{Method}       & \textbf{Ref.}    & \textbf{Exist↑}  & \textbf{Object↑} & \textbf{Status↑} & \textbf{Comparison} \\
\toprule
\multirow{4}{*}{NusceneQA Dataset} & EM-VLM4AD\textsubscript{Base}  & {CVPR’24} & {75.9}   & {40.2}   & {51.9}    & {63.2}    \\
                                   & EM-VLM4AD\textsubscript{Large} & {CVPR’24} & {70.2}   & {38.6}   & {45.7}    & {60.9}    \\
                                   & NusceneQA-Agent             & {AAAI’24} & \textbf{84.8}   & {52.3}   & {59.8}    & {70.0}   \\
                                   & LaVida Drive (Ours)         & {-}       & \underline{78.0}    & \textbf{52.8}    & \textbf{60.9}     & \textbf{70.5}   \\
\bottomrule
\end{tabular}
}
\label{main}
\end{table*}

\begin{table}[ht]
\caption{Ablation study on select ratio and compress ratio while fixing the overall reduction ratio to 168. Bold indicates the highest value.}

\centering
\renewcommand{\arraystretch}{3.5}  
\resizebox{\columnwidth}{!}{  
\begin{tabular}{c c c c c c} 
  \toprule[1.0mm]
  \textbf{\fontsize{28}{35}\selectfont\makecell{Select\\Ratio}} & \textbf{\fontsize{28}{35}\selectfont\makecell{Compress\\Ratio}} & \textbf{\fontsize{28}{40}\selectfont BLEU-4} & \textbf{\fontsize{28}{40}\selectfont METEOR} & \textbf{\fontsize{28}{40}\selectfont ROUGE-L} & \textbf{\fontsize{28}{40}\selectfont CIDEr} \\
  \midrule[0.5mm]
  \fontsize{24}{20}\selectfont - & \fontsize{28}{20}\selectfont 168 & \fontsize{28}{20}\selectfont 48.0 & \fontsize{28}{20}\selectfont 34.5 & \fontsize{28}{20}\selectfont 72.0 & \fontsize{28}{20}\selectfont 3.20 \\
  \fontsize{24}{20}\selectfont 2 & \fontsize{28}{20}\selectfont 84 & \fontsize{28}{26}\selectfont \textbf{51.3} & \fontsize{28}{26}\selectfont \textbf{38.0} & \fontsize{28}{26}\selectfont \textbf{73.9} & \fontsize{28}{26}\selectfont \textbf{3.32} \\
  \fontsize{24}{26}\selectfont 3 & \fontsize{28}{26}\selectfont 56 & \fontsize{28}{26}\selectfont 50.7 & \fontsize{28}{26}\selectfont 37.8 & \fontsize{28}{26}\selectfont 73.3 & \fontsize{28}{26}\selectfont 3.22 \\
  \fontsize{24}{26}\selectfont 6 & \fontsize{28}{26}\selectfont 26 & \fontsize{28}{26}\selectfont 49.4 & \fontsize{28}{26}\selectfont 36.6 & \fontsize{28}{26}\selectfont 73.0 & \fontsize{28}{26}\selectfont 3.26 \\
  \fontsize{24}{26}\selectfont 84 & \fontsize{28}{26}\selectfont 2 & \fontsize{28}{26}\selectfont 46.3 & \fontsize{28}{26}\selectfont 34.1 & \fontsize{28}{26}\selectfont 70.5 & \fontsize{28}{26}\selectfont 3.13 \\
  \fontsize{24}{26}\selectfont 168 & \fontsize{28}{26}\selectfont - & \fontsize{28}{26}\selectfont 42.0 & \fontsize{28}{26}\selectfont 32.8 & \fontsize{28}{26}\selectfont 63.5 & \fontsize{28}{26}\selectfont 2.96 \\
  \bottomrule[1.0mm]

\end{tabular}
}
\label{table:select_ratio}
\end{table}

\begin{table*}[ht]
\caption{Component-wise ablation experiment results. \textbf{\textcolor{rgb:red,0;green,0.5;blue,0.2}{+}} denotes an added module or method, and \textcolor{cyan}{©} denotes a changed module or method.}

\centering
\setlength{\tabcolsep}{5mm}
\begin{tabular}{l|c|cccc}
\toprule
\textbf{Method} & \#Tokens & \textbf{BLEU-4} & \textbf{METEOR} & \textbf{ROUGE-L}& \textbf{CIDEr} \\ \midrule
Baseline         & 49*6     & 45.4           & 34.5 & 72.0          & 3.20          \\ 
\textbf{\textcolor{rgb:red,0;green,0.5;blue,0.2}{+}} High Solution Patch &49*6*28     & 49.4 $\uparrow$4.0 & 37.0 $\uparrow$2.5 & 73.7 $\uparrow$1.7 &3.29$\uparrow$0.9 \\ 
\textbf{\textcolor{rgb:red,0;green,0.5;blue,0.2}{+}} MLP+Pooling    & 49    & 48.0 $\downarrow$1.4  & 36.5 $\downarrow$0.5 & 72.3 $\downarrow$1.4   &3.20$\downarrow$0.9     \\ 
\textbf{\textcolor{cyan}{©}} Token Selection & 49     & 50.7 $\uparrow$2.7 & 37.0 $\uparrow$0.5 & 73.0 $\uparrow$0.7 & 3.26 $\uparrow$0.6\\  
\textbf{\textcolor{rgb:yellow,0;green,0.5;blue,0.2}{+}} Token Enhancement & 49     & \textbf{51.3} $\uparrow$0.6 & \textbf{38.0} $\uparrow$1.0 &  \textbf{73.9}$\uparrow$0.9 & \textbf{3.32} $\uparrow$0.6\\ 
\bottomrule
\end{tabular}
\label{tab:component}
\end{table*}

\begin{table}[ht]
\caption{Ablation among input type. We compare the response performance of different types of the dataset on the same test models. Bold indicates the highest value.}
\centering
\renewcommand{\arraystretch}{4.0}  
\resizebox{\columnwidth}{!}{  
\begin{tabular}{c c c c c c} 
  \toprule[1.0mm]
  \textbf{\fontsize{28}{26}\selectfont Multiview} & \textbf{\fontsize{28}{26}\selectfont Multiframe} & \textbf{\fontsize{28}{26}\selectfont BLEU-4} & \textbf{\fontsize{28}{26}\selectfont METEOR} & \textbf{\fontsize{28}{26}\selectfont ROUGE-L} & \textbf{\fontsize{28}{26}\selectfont CIDEr} \\
  \midrule[0.5mm]
  \fontsize{28}{26}\selectfont \checkmark & \fontsize{28}{26}\selectfont - & \fontsize{28}{26}\selectfont 48.30 & \fontsize{28}{26}\selectfont 36.5 & \fontsize{28}{26}\selectfont 72.3 & \fontsize{28}{26}\selectfont 3.23 \\
  \fontsize{28}{26}\selectfont - & \fontsize{28}{26}\selectfont \checkmark & \fontsize{28}{26}\selectfont 50.8 & \fontsize{28}{26}\selectfont 37.2 & \fontsize{28}{26}\selectfont \textbf{74.0} & \fontsize{28}{26}\selectfont \textbf{3.32} \\
  \fontsize{28}{26}\selectfont \checkmark & \fontsize{28}{26}\selectfont \checkmark & \fontsize{28}{26}\selectfont \textbf{51.3} & \fontsize{28}{26}\selectfont \textbf{38.0} & \fontsize{28}{26}\selectfont 73.9 & \fontsize{28}{26}\selectfont \textbf{3.32} \\
  \midrule[1.0mm]
\end{tabular}
}
\label{table:input_type}
\end{table}

In this section, we conduct extensive experiments on Lavida Drive and analyze the results, including both quantitative and qualitative evaluations. Finally, we perform ablation studies to validate the effectiveness of each module.

\subsection{Setup}
\noindent\textbf{Dataset:} Following EM-VLM4AD~\cite{gopalkrishnan2024multiframelightweightefficient} protocols, we use identical training, validation, and test splits on the DriveLM~\cite{sima2023drivelm} dataset, which covers tasks like perception, prediction, and decision-making to evaluate generalizability. Additionally, we test on the NusceneQA~\cite{qian2024nuscenesqamultimodalvisualquestion} dataset for fine-grained perception, comparing to traditional detection-based methods. The DriveLM training set includes approximately 340,184 unique multi-view QA pairs, with 18,899 pairs for testing and validation. The NusceneQA training and test sets contain 459,941 and 83,337 question-answer pairs, respectively.


\noindent\textbf{Metrics:} To ensure fairness and reproducibility, when evaluating the DriveLM dataset, we use the same metrics as EM-VLM4AD, assessing model performance from four perspectives: BLEU-4~\cite{papineni2002bleu}, ROUGE-L~\cite{lin2004rouge}, METEOR \cite{banerjee2005meteor}, and CIDEr~\cite{vedantam2015cider}. For the NusceneQA dataset, we align with the metrics proposed by the dataset's authors, evaluating accuracy across four query format categories: Exist, Object, Status, and Comparison.

\noindent\textbf{Models:} We employ the CLIP text and vision encoders as our text and image encoders, respectively, and utilize TimeSformer\cite{gberta_2021_ICML} as a video encoder for multi-frame input processing. The base language model used is T5-medium\cite{2020t5}.

\noindent\textbf{Implementation Details:} Each model is trained on a single NVIDIA A100 Tensor Core GPU. The image encoder, text encoder, and video encoder are frozen, while other parameters, including those in larger models, are trained with an initial learning rate of 1e-4 and a weight decay of 0.05. The batch size is set to 4. Each model is trained for 12 epochs on the training set, with each image divided into 4x7 patches of size 224x224.

\subsection{Overall Performance}
\noindent\textbf{Quantitative Comparison:} 
In Tab.~\ref{main}, we first compare our model with prior works on the DriveLM dataset, including EM-VLM4AD~\cite{gopalkrishnan2024multiframelightweightefficient} and DriveLM-Agent~\cite{sima2023drivelm}. Lavida Drive outperforms our baseline method in overall performance metrics, although DriveLM-Agent slightly exceeds our model in BLEU-4 score. However, it is important to note that DriveLM-Agent has a significantly larger parameter count, reaching 3.96B. Next, we fine-tune the pretrained model on the NusceneQA training set and evaluate it on the test set. Without using 3D detector outputs and prompts, our model achieves competitive results.

\noindent\textbf{Qualitative Comparison:}
As shown in Fig.~\ref{fig:4}, we present LaVida Drive's performance across various tasks. To analyze its dynamic perception of multi-view image inputs based on textual queries, we visualize the tokens selected by LaVida Drive in different scenarios. The analysis demonstrates that the model adaptively selects the most relevant tokens guided by keywords such as "car" and "pedestrians." This process resembles human reasoning during driving, where relevant information is first filtered before drawing inferences. This dynamic token selection prior to model input underscores the system's reliability and interpretability.
\subsection{Ablation Studies}
To verify the effectiveness of each module in our model, we designed a series of ablation studies, focusing on the selection and compression ratios, model components, and input types.

\noindent\textbf{Select and Compress Ratio Ablation:}
We fixed the overall compression ratio at 168 and tested various combinations of the selection and MLP compression ratios to assess their impact on model performance. As shown in Tab.~\ref{table:select_ratio}, initially, we compressed tokens using MLP alone, without selection, which gave suboptimal results. Then, we gradually increased the selection ratio while decreasing the compression ratio, observing peak performance at a selection ratio of 2, after which performance declined. Finally, we reduced tokens by a factor of 168 using selection alone (without MLP compression), resulting in a significant performance drop.

These results indicate that excessive redundancy and irrelevant information hinder VLM training and inference. However, overly reducing token selection leads to the loss of valuable information. Therefore, balancing the selection factor and MLP compression ratio is crucial for achieving higher model performance with fewer tokens.

\noindent\textbf{Components-wise Ablation:} 
Tab.~\ref{tab:component} presents the component-wise experimental results of our method. We start by using the 224×224 downsampled feature map from the image encoder as low-resolution visual embeddings, producing 49×6 visual tokens, which we use as the baseline. Next, we replace simple downsampling with high-resolution 224×224 patches. This approach improves performance by +4.0\%, +2.5\%, +1.7\%, and +0.9\%, but increases token overhead. To address this issue, we first apply an MLP layer to downsample the tokens, reducing their count to 49; however, this results in performance declines of -1.4\ -0.5\%, -1.4\%, and -0.9\%. By replacing the MLP layer with our proposed Token Selection module, we effectively identify the most relevant tokens, leading to performance gains of +2.7\%, +0.5\%, +0.7\%, and +0.6\%, Finally, incorporating Token Recovery and Token Enhancement methods leads to further improvements of +0.6\%, +1.0\%, +0.9\%, and +0.6\%, respectively while maintaining the token count.

\noindent\textbf{Input Type Ablation:} 
To validate the robustness of our algorithm under scenarios where input data is lost in complex or adverse conditions, we tested different input configurations: single-frame multi-view, multi-frame single-view, and single-frame single-view. By analyzing the model's performance after removing parts of the complete multi-view multi-frame dataset, we assess the model's robustness and the impact of multi-frame and multi-view data on its performance. As shown in Tab.~\ref{table:input_type}, our model maintains relatively good performance even after some data is removed. Although there is a significant collapse in performance when using the single-frame single-view dataset, it still outperforms our baseline.

\section{Conclusion}

In this work, we present LaVida Drive, a novel framework for VQA in autonomous driving, which effectively integrates high-resolution spatial perception with temporal dynamics. 
By leveraging query-aware token selection and spatial-temporal token enhancement, our approach reduces computational overhead without sacrificing fine-grained visual detail, enabling more efficient inference through selective processing of relevant detail visual cues and ensuring coherent information flow across frames. LaVida Drive provides a promising framework for real-time VQA systems in autonomous driving, balancing computational efficiency with detailed perception. It effectively integrates spatial and temporal information, laying the groundwork for intelligent systems that can handle complex, dynamic driving environments.

{
    \small
    \bibliographystyle{ieeenat_fullname}
    \bibliography{main}
}


\end{document}